\pdfoutput=1

\documentclass[11pt]{article}

\usepackage[]{EACL2023}

\usepackage{times}
\usepackage{latexsym}
\usepackage{booktabs}
\usepackage{multirow}
\usepackage{tabularx}
\usepackage{xspace}
\usepackage{xcolor,colortbl}
\usepackage{soul}
\usepackage{enumitem}
\usepackage{graphicx}
\usepackage{amsmath}
\usepackage{subcaption}
\usepackage{todonotes}
\usepackage{siunitx}
\usepackage[T1]{fontenc}

\usepackage[utf8]{inputenc}

\usepackage{microtype}

\usepackage{inconsolata}
\definecolor{Gray}{gray}{0.85}
\newcolumntype{g}{>{\columncolor{Gray}}c}

\title{%
Can Demographic Factors Improve Text Classification?\\%
Revisiting Demographic Adaptation in the Age of Transformers%
}

\author{Chia-Chien Hung\textsuperscript{1,5}, Anne Lauscher\textsuperscript{2}, Dirk Hovy\textsuperscript{3}, \\\textbf{Simone Paolo Ponzetto}\textsuperscript{1} and \textbf{Goran Glava\v{s}}\textsuperscript{4} \\
  \textsuperscript{1}Data and Web Science Group, University of Mannheim, Germany \\
  \textsuperscript{2}Data Science Group, University of Hamburg, Germany \\
  \textsuperscript{3}MilaNLP, Bocconi University, Italy \quad
  \textsuperscript{4}CAIDAS, University of Würzburg, Germany \\
  \textsuperscript{5}NEC Laboratories Europe GmbH, Heidelberg, Germany \\
  \texttt{\{chia-chien.hung, ponzetto\}@uni-mannheim.de}\\ \texttt{anne.lauscher@uni-hamburg.de}, \texttt{dirk.hovy@unibocconi.it}\\ \texttt{goran.glavas@uni-wuerzburg.de} \\}

\begin{document}
\maketitle
\begin{abstract}
Demographic factors (e.g., gender or age) shape our language. 
Previous work showed that incorporating demographic factors can consistently improve performance for various NLP tasks with traditional NLP models. In this work, we investigate whether these previous findings still hold with state-of-the-art pretrained Transformer-based language models (PLMs). We use three common specialization methods proven effective for incorporating external knowledge into pretrained Transformers (e.g., domain-specific or geographic knowledge). We adapt the language representations for the demographic dimensions of gender and age, using continuous language modeling and dynamic multi-task learning for adaptation, where we couple language modeling objectives with the prediction of demographic classes. 
Our results, when employing a multilingual PLM, show substantial gains in task performance across four languages (English, German, French, and Danish), which is consistent with the results of previous work. 
However, controlling for confounding factors -- primarily domain and language proficiency of Transformer-based PLMs -- shows that downstream performance gains from our demographic adaptation do \textit{not} actually stem from demographic knowledge. Our results indicate that demographic specialization of PLMs, while holding promise for positive societal impact, still represents an unsolved problem for (modern) NLP.

\end{abstract}


\newcommand{\al}[1]{\textcolor{purple}{#1}}
\section{Introduction}
\label{sec:intro}
Demographic factors like social class, education, income, age, or gender, categorize people into specific groups or populations. At the same time, demographic factors both shape and are reflected in our language \cite[e.g.,][]{trudgill, eckert}.
%
A large body of work focused on modeling demographic language variation, especially the correlations between words and demographic factors~\cite[\emph{inter alia}]{bamman-etal-2014-distributed, garimella-etal-2017-demographic, welch-etal-2020-compositional}. In a similar vein, \newcite{volkova-etal-2013-exploring} and \newcite{hovy-2015-demographic} demonstrated that explicitly incorporating demographic information in language representations improves performance on downstream NLP tasks, e.g., topic classification or sentiment analysis. 
However, these observations rely on approaches that leverage gender-specific lexica to specialize word embeddings and text encoders (e.g., recurrent networks) that have not been pretrained for (general purpose) language understanding.  
To date, the benefits of demographic specialization have not been tested with Transformer-based~\citep{Transformer} pretrained language models (PLMs), which have been shown to excel on the vast majority of NLP tasks and even surpass human performance in some cases~\citep{wang-etal-2018-glue}. 

More recent studies focus mainly on monolingual English datasets and introduce demographic features in task-specific fine-tuning~\cite{voigt-etal-2018-rtgender, buechel-etal-2018-modeling}, which limits the benefits of demographic knowledge to tasks at hand. In this work, we investigate the (task-agnostic) demographic specialization of PLMs, aiming to impart the associations between demographic categories and linguistic phenomena into the PLMs parameters. If successful, such specialization could benefit any downstream NLP task in which demographic factors (i.e., demographically conditioned language phenomena) matter. 
For this, we adopt intermediate training paradigms that have been proven effective for the specialization of PLMs for other types of knowledge, e.g., in domain, language, and geographic  adaptation~\citep{glavas-etal-2020-xhate, hung-etal-2022-ds, hofmann2022geographic}.
To this effect, we perform (i) continued language modeling on text corpora produced by a demographic group and (ii) dynamic multi-task learning~\citep{kendall2018multi}, wherein we combine language modeling with the prediction of demographic categories. 

We evaluate the effectiveness of the demographic PLM specialization on both intrinsic (demographic category prediction) and extrinsic (sentiment classification and topic detection) evaluation tasks across four languages: English, German, French, and Danish, using a multilingual corpus of reviews~\citep{hovy2015user} annotated with demographic information. In line with earlier findings \cite{hovy-2015-demographic}, our initial experiments based on a multilingual PLM~\citep[mBERT;][]{devlin-etal-2019-bert}, render demographic specialization effective: we observe gains in most tasks and settings. 
Through a set of controlled experiments, where we (1) adapt with in-domain language modeling alone, without leveraging demographic information, (2) demographically specialize \textit{monolingual} PLMs of evaluation languages, (3) carry out a meta-regression analysis over dimensions that drive the performance, and (4) analyze the topology of the representation spaces of demographically specialized PLMs, we show, however, that most of the original gains can be attributed to confounding effects of language and/or domain specialization.

Our findings indicate that specialization approaches, proven effective for other types of knowledge, fail to adequately instill demographic knowledge into PLMs, making demographic specialization of NLP models an open problem in the age of large pretrained Transformers. Our research code is
publicly available at: \url{https://github.com/umanlp/SocioAdapt}.  




\section{Demographic Adaptation}
Our goal is to inject demographic knowledge through intermediate PLM training in a task-agnostic manner. To achieve this goal, we train the PLM in a dynamic multi-task learning setup~\citep{kendall2018multi}, in which we couple masked language modeling (MLM-ing) with predicting the demographic category -- gender or age group of the text author. Such multi-task learning setup is designed to force the PLM to learn associations between the language constructs and demographic groups, if these associations are salient in the training corpora. 

\paragraph{Masked Language Modeling (MLM).}
\label{ss:mlm}
Following successful work on pretraining via language modeling for domain-adaptation~\citep{gururangan-etal-2020-dont, hung-etal-2022-ds}, 
we investigate the effect of running standard MLM-ing on the 
text corpora of a specific demographic dimension (e.g., gender-related corpora). 
We compute the MLM loss $L_{mlm}$ in the common way, as negative log-likelihood of the true token probability.


%

\paragraph{Demographic Category Prediction.}
\label{ss:socio_predict}

In the multi-task learning setup, the representation of the input text, as output by the Transformer, is additionally fed into a classification head that predicts the corresponding demographic category: \textit{age} (below 35 and above 45\footnote{As suggested by \citet{hovy-2015-demographic} the split for the age ranges result in roughly equally-sized data sets for each sub-group and is non-contiguous, avoiding fuzzy boundaries.}), and  \textit{gender} (female and male). The demographic prediction loss $L_{dem}$ is computed as the standard binary cross-entropy loss. 

We experiment with two different ways of predicting the demographic category of the text:   
(i)~from the transformed representation of the sequence start token (\texttt{[CLS]}) and (ii)~from the contextualized representations of each masked token. We hypothesized that the former variant, in which we predict the demographic class from the \texttt{[CLS]} token representation, would establish links between more complex demographically condition linguistic phenomena (e.g., syntactic patterns or patterns of compositional semantics that a demographic group might exhibit), whereas the latter -- predicting demographic class from representations of masked tokens -- is more likely to establish simpler lexical links, i.e., capture the vocabulary differences between the demographic groups. 

\paragraph{Multi-Task Learning.}
\label{ss:mtl}
Since both losses can be computed from the same input instances, we opt for joint multi-task learning (MTL)
and resort to dynamic MTL based on the \emph{homoscedastic} uncertainty of the losses, wherein the loss variances are used to balance the contributions of the tasks~\citep{kendall2018multi}. The intuition is that more effective MTL occurs if we dynamically assign less importance to more uncertain tasks, as opposed to assigning uniform task weights throughout the whole training. Homoscedastic uncertainty weighting in MTL has been effective in different NLP settings~\citep{lauscher-etal-2018-investigating, hofmann2022geographic}. 
In our scenario, $L_{mlm}$ and $L_{dem}$ are measured on different scales in which the model would favor (i.e., be more confident for) one objective than the other. The confidence level of the model prediction for each task would change throughout the training progress: this makes dynamic weighting desirable. 
We dynamically prioritize the tasks via homoscedastic uncertainties $\sigma_t$: 

\small{
\begin{equation}
\tilde{L}_t = \frac{1}{2\sigma_t^{2}}L_t + \log \sigma_t\,,
\end{equation}
}

\normalsize
\noindent where $\sigma_t^2$ is the variance of the task-specific loss over training instances for quantifying the uncertainty of the task $t\in\{mlm, dem\}$. In practice, we train the network to predict the log variance, $\eta_t :=\log \sigma_t^2$, since it is more numerically stable than regressing the variance $\sigma_t^2$, as the log avoids divisions by zero. The adjusted losses are then computed as:

\small{
\begin{equation}
\tilde{L}_t = \frac{1}{2}(e^{-\eta_t}L_t + \eta_t)\,.
\end{equation}
}

\normalsize
\noindent The final loss we minimize is the sum of the two uncertainty-adjusted losses: $\tilde{L}_{mlm}$ + $\tilde{L}_{dem}$.

\section{Experimental Setup}
\label{s:setup}
\setlength{\tabcolsep}{8.0pt} 
\label{sec:method}

\begin{table*}[t]
\centering
\scriptsize{
\begin{tabular}{llrrrrrrrr}
\toprule
\multicolumn{1}{l}{} & \multicolumn{1}{l}{} & \multicolumn{4}{c}{\textit{\textbf{gender}}}                                              & \multicolumn{4}{c}{\textit{\textbf{age}}}                                                 \\\cmidrule(lr){3-6}\cmidrule(lr){7-10}
\textbf{Country}     & \textbf{Language}    & \multicolumn{2}{c}{\textbf{Specialization}} & \textbf{SA, AC-SA} & \textbf{TD, AC-TD} & \multicolumn{2}{c}{\textbf{Specialization}} & \textbf{SA, AC-SA} & \textbf{TD, AC-TD} \\ \cmidrule(lr){3-4}\cmidrule(lr){5-6}\cmidrule(lr){7-8}\cmidrule(lr){9-10}
                     &                      & \multicolumn{1}{c}{\textbf{F}}         & \multicolumn{1}{c}{\textbf{M}}        & \multicolumn{2}{c}{\textbf{F / M}}      & \multicolumn{1}{c}{\textbf{<35}}         & \multicolumn{1}{c}{\textbf{>45}}        & \multicolumn{2}{c}{\textbf{<35 / >45}}     \\ \midrule
Denmark              & Danish               & 1,596,816            & 2,022,349           & 250,485             & 120,805             & 833,657             & 494,905            & 75,300              & 44,815              \\
France               & French               & 489,778             & 614,495            & 67,305              & 55,570              & 40,448              & 36,182             & 6,570               & 6,120               \\
Germany              & German               & 210,718             & 284,399            & 28,920              & 30,580              & 66,342              & 47,308             & 5,865               & 8,040               \\
UK                   & English              & 1,665,167            & 1,632,894           & 156,630             & 183,995             & 231,905             & 274,528            & 26,325              & 22,095              \\
US                   & English              & 575,951             & 778,877            & 72,270              & 61,585              & 124,924             & 70,015             & 6,495               & 12,090  \\    \bottomrule       
\end{tabular}%
}
\vspace{-0.5em}
\caption{Number of instances in different portions of the Trustpilot dataset~\citep{hovy2015user} used in our experiments. For each country (Denmark, France, Germany, UK, and US), we report the size of the specialization and fine-tuning portions, the latter for each of the two extrinsic tasks: Sentiment Analysis (SA) and Topic Detection (TD). Note that we use the same SA and TD reviews for the intrinsic AC tasks of predicting the demographic categories (denoted AC-SA and AC-TD, respectively). Numbers are shown separately for the two demographic dimensions: gender and age. For fine-tuning datasets (for SA/AC-SA, and for TD/AC-TD), we indicate the number of instances in each category (which is the same for both categories: F and M for gender, <35 and >45 for age). We split the fine-tuning datasets randomly into train, validation, and test portions in the 60/20/20 ratio.    
}
\vspace{-0.5em}
\label{tab:dataset_info}
\end{table*}

\label{sec:experiments}
Here we describe evaluation tasks and provide details on the data used for demographic specialization and downstream evaluation.
\paragraph{Evaluation Tasks.}
We follow~\citet{hovy-2015-demographic} and measure the effects of demographic specialization of PLMs on three text-classification tasks, coupling intrinsic demographic \textit{attribute classification} (\textbf{AC}) with two extrinsic text classification tasks: \textit{sentiment analysis} (\textbf{SA}) and \textit{topic detection} (\textbf{TD}). As an intrinsic evaluation task, AC directly tests if the intermediate demographic specialization results in a PLM that can be more effectively fine-tuned to predict the same demographic classes used in the intermediate specialization: PLMs (vanilla PLM and our demographically specialized counterpart) -- are fine-tuned in a supervised fashion to predict the demographic class (gender or age) of the text author. SA is a ternary classification task in which the reviews with ratings of $1$, $3$, and $5$ stars represent instances of \textit{negative}, \textit{neutral}, and \textit{positive} class, respectively. TD classifies texts into 5 different topic categories. We report the $F_{1}$-measure for each task following \citet{hovy-2015-demographic}.

\paragraph{Data.}
\label{ssec:data}
We carry out our core experimentation on the multilingual demographically labeled dataset of reviews~\citep{hovy2015user}, created from the internationally popular user review website Trustpilot.\footnote{\url{https://www.trustpilot.com/}} For comparison and consistency, we work with exactly the same data portions as~\citet{hovy-2015-demographic}: collections that cover (1) two most prominent demographic dimensions -- \emph{gender} and \emph{age}, with two categories in each 
(gender: male or female; age: below 35 or above 45\footnote{As suggested by \citet{hovy-2015-demographic}, the split for the age ranges results in roughly equally-sized data sets for each sub-group and is non-contiguous, avoiding fuzzy boundaries.}) and (2) five countries (four languages): United States (US), Denmark, Germany, France, and United Kingdom (UK).

To avoid any information leakage, we ensure -- for each country-demographic dimension collection (e.g., US, gender) -- that there is zero overlap between the portions we select for intermediate demographic specialization and portions used for downstream fine-tuning and evaluation (for AC, SA, and TD).   
(Specialization). 
For TD, we aim to eliminate the confounding effect of demographically-conditioned label distributions (e.g., female authors wrote reviews for \textit{clothing store} more frequently than male authors; vice-versa for \textit{electronics \& technology}). To this effect,  
we select, for each country, reviews from the five most frequent topics and sample the same number of reviews in each topic for both demographic groups (i.e., \textit{male} and \textit{female} for gender; \textit{below 35} and \textit{above 45} for age).  
For the intrinsic AC task (i.e., fine-tuning to predict either gender or age category), we report the results for two different review collections: the first is the set of reviews that have, besides the demographic classes, been annotated with sentiment labels (we refer to this as AC-SA) and the second are the reviews that have topic labels (i.e., product/service category; we refer to this portion as AC-TD). For these fine-tuning and evaluation datasets, we make sure that the two demographic classes (\textit{male} and \textit{female} for gender \textit{under 35} and \textit{above 45} for age) are equally represented in each dataset portion (train, development, and test).   
Table~\ref{tab:dataset_info} displays the numbers of reviews for each country, demographic aspect, and dataset portion (specialization vs. fine-tuning). 


For intermediate specialization of the multilingual model, we randomly sample 100K instances per demographic group from the \textit{gender} specialization portion and 50K instances each from the texts reserved for \textit{age} specialization concatenated across all 5 countries. For the specialization of monolingual PLMs, we randomly sample 
the same number of instances but from the specialization portions of a \textit{single} country. Following the established procedure~\citep[e.g.,][]{devlin-etal-2019-bert, liu2019roberta}, we dynamically mask 15\% of the tokens in the demographic specialization portions for MLM.

\paragraph{Pre-trained language models.}
\label{subsec:baselines}
 Given that we experiment with Trustpilot data in four different languages, in our core experiments, we resorted to multilingual BERT (mBERT)\footnote{We load the \texttt{bert-base-multilingual-cased} weights from HuggingFace Transformers.}~\citep{devlin-etal-2019-bert} as the starting PLM. This allows us to merge the (fairly large) specialization portions of Trustpilot in different languages (see Table \ref{tab:dataset_info}) and run a single multilingual demographic specialization procedure on the combined multilingual review corpus.
 We then fine-tune the demographically-specialized mBERT and evaluate downstream task performance separately for each of the five countries (using train, development, and test portions of the respective country). 
 We report the results for two different variants of our dynamic multi-task demographic specialization (DS): (1) when the demographic category is predicted from representations of masked tokens (DS-Tok) and (2) when we predict the demographic category from the encoding of the whole sequence (i.e., review; this version is denoted with DS-Seq). We compare these demographic-specialized PLM variants against two baselines: vanilla PLM and PLM specialized on the same review corpora as our MTL variants but only via MLM-ing (i.e., without providing the demographic signal).
 

\paragraph{Training and Optimization.}
In demographic specialization training, we fix the maximum sequence length to $128$ subword tokens. We train for $30$ epochs in batches of $32$ instances and search for the optimal learning rate among the following values: $\{5\cdot 10^{-5}, 1\cdot 10^{-5}, 1\cdot 10^{-6}\}$. We apply early stopping based on the development set performance: we stop if the joint MTL loss does not improve for~3~epochs). 
%
%
For downstream fine-tuning and evaluation, we train for maximum $20$ epochs in batches of $32$. We search for the optimal learning rate between the following values: $\{5\cdot 10^{-5}, 1\cdot 10^{-5}, 5\cdot 10^{-6}, 1\cdot 10^{-6}\}$ and apply early stopping based on the validation set performance (patience: 5 epochs). We use AdamW~\citep{loshchilov2018decoupled} as the optimization algorithm.

\section{Results and Discussion}
\label{sec:results}



We first discuss the results of multilingual demographic specialization with mBERT as the PLM ~(\S\ref{subsec:initial_results}). We then provide a series of control experiments in which we isolate the effects that contribute to performance gains of demographically specialized PLMs (\S\ref{ssec:control_results}). 

\setlength{\tabcolsep}{4.6pt}
\begin{table*}[t!]
\centering
\scriptsize{
\begin{tabular}{cl cc ccc ccc || cc ccc ccc }
\toprule
& & \multicolumn{8}{c}{\textit{Demographic: \textbf{gender}}} & \multicolumn{8}{c}{\textit{Demographic: \textbf{age}}} \\ \cmidrule(lr){3-10} \cmidrule(lr){11-18}
& & \multicolumn{2}{c}{\textbf{Gender~class.}} & \multicolumn{3}{c}{\textbf{SA}} & \multicolumn{3}{c}{\textbf{TD}} & \multicolumn{2}{c}{\textbf{Age~class.}} & \multicolumn{3}{c}{\textbf{SA}} & \multicolumn{3}{c}{\textbf{TD}} \\ \cmidrule(lr){3-4} \cmidrule(lr){5-7}\cmidrule(lr){8-10}\cmidrule(lr){11-12} \cmidrule(lr){13-15}\cmidrule(lr){16-18}
\textbf{Country} &  \textbf{Model} & \textbf{AC-SA} & \textbf{AC-TD} & F & M & X & F & M & X & \textbf{AC-SA} & \textbf{AC-TD} & <35 & >45 & X & <35 & >45 & X \\ \midrule
\multirow{4}{*}{\textbf{Denmark}} & mBERT & 64.0 & 61.8 & 69.2 & 64.8 & 67.2 & 59.3 & 58.3 & 59.0 
& 57.2 & 64.5 & 62.7 & 62.7 & 62.9 & 56.1 & 52.2 & 53.4 \\ 
& MLM & \textbf{65.2} & 63.4 & 69.5 & \textbf{65.8} & 67.8 & 59.7 & 58.8 & \textbf{59.4} 
& \textbf{65.5} & 65.1 & 63.3 & 62.1 & 63.0 & \textbf{57.1} & 52.6 & 54.1 \\ \cmidrule(lr){2-18}
& DS-Seq & 64.9 & 63.5 & \textbf{69.9} & 65.7 & 67.7 & 59.7 & 57.8 & 59.1 
& 65.2 & \textbf{65.2} & 63.1 & 62.9 & 63.0 & 56.9 & \textbf{53.3} & \textbf{54.5} \\
& DS-Tok & 65.0 & \textbf{63.5} & 69.1 & 65.6 & \textbf{68.0} & \textbf{59.9} & \textbf{58.9} & 59.0 
& 65.3 & 64.6 & \textbf{64.2} & \textbf{63.3} & \textbf{63.2} & 56.2 & 53.2 & 54.3 \\\midrule
\multirow{4}{*}{\textbf{Germany}} & mBERT & 59.5 & 57.9 & 66.1 & 63.2 & 64.5 & 67.8 & 65.6 & 65.8 
& 58.0 & 56.9 & 52.6 & 55.0 & 55.0 & 60.1 & 55.3 & 57.1\\
 & MLM & 61.2 & 60.1 & \textbf{67.7} & \textbf{65.3} & 66.1 & \textbf{68.6} & 67.0 & \textbf{67.1} 
& \textbf{61.1} & \textbf{58.9} & 53.6 & 55.5 & 56.7 & \textbf{61.5} & 56.5 & 58.7 \\ \cmidrule(lr){2-18}
 & DS-Seq & 60.1 & \textbf{60.3} & 66.7 & 64.0 & 65.7 & 67.6 & 65.7 & 66.4 
& 56.4 & 58.2 & \textbf{53.8} & 55.3 & 55.5 & 60.8 & \textbf{57.6} & \textbf{59.3} \\
 & DS-Tok & \textbf{62.9} & 58.3 & 66.8 & 64.3 & \textbf{66.8} & 68.3 & \textbf{67.0} & 66.7 
& 56.6 & 57.4 & 53.0 & \textbf{56.5} & \textbf{56.7} & 59.3 & 56.5 & 59.3 \\\midrule
\multirow{4}{*}{\textbf{US}} & mBERT & 62.6 & 58.1 & 66.3 & 64.4 & 66.0 & 71.2 & 68.4 & 70.2 
& 62.9 & 60.7 & 57.7 & 57.9 & 57.8 & 68.0 & 64.3 & 64.3 \\
 & MLM & 63.3 & \textbf{59.6} & 67.3 & 66.2 & 66.9 & 72.1 & 69.4 & 70.3 
& \textbf{63.6} & \textbf{61.9} & 59.4 & 57.8 & \textbf{58.2} & 69.0 & 64.2 & 65.2 \\ \cmidrule(lr){2-18}
 & DS-Seq & \textbf{63.8} & 59.2 & 67.2 & 66.3 & 67.0 & 72.3 & 69.2 & 70.4 
& 60.7 & 61.5 & 59.3 & 57.9 & 58.0 & \textbf{69.8} & 64.4 & \textbf{65.8} \\
 & DS-Tok & 62.2 & 58.8 & \textbf{68.0} & \textbf{66.4} & \textbf{67.3} & \textbf{72.8} & \textbf{69.5} & \textbf{70.5} 
& 59.7 & 61.2 & \textbf{59.9} & \textbf{58.6} & 57.8 & 69.2 & \textbf{65.4} & 64.9 \\ \midrule
\multirow{4}{*}{\textbf{UK}} & mBERT & 61.9 & 63.1 & 71.0 & 69.0 & 69.7 & 70.4 & 67.9 & 68.9 
& 65.1 & 65.2 & 63.8 & 63.9 & 63.7 & 64.7 & 67.1 & 66.3 \\
 & MLM & 63.0 & 65.3 & 72.0 & 70.4 & 71.0 & 70.6 & 67.9 & 69.8 
& \textbf{65.4} & \textbf{65.6} & 62.8 & 62.0 & 63.0 & 65.1 & 67.3 & 67.3 \\ \cmidrule(lr){2-18}
 & DS-Seq & 63.4 & 64.9 & 72.9 & 70.9 & 71.7 & 70.6 & 68.2 & 69.8 
& 65.3 & 62.8 & 63.8 & 64.9 & 64.9 & 66.0 & \textbf{68.1} & 66.5 \\
 & DS-Tok & \textbf{63.5} & \textbf{65.6} & \textbf{73.0} & \textbf{71.0} & \textbf{71.9} & \textbf{70.8} & \textbf{68.2} & \textbf{69.9} 
 & 64.0 & 62.8 & \textbf{64.6} & \textbf{65.2} & \textbf{65.1} & \textbf{66.4} & 67.3 & \textbf{67.6} \\\midrule
\multirow{4}{*}{\textbf{France}} & mBERT & 63.9 & 61.2 & 69.3 & 67.0 & 67.8 & 44.6 & 42.4 & 43.1 
& 55.7 & 56.6 & 59.6 & 57.4 & 61.5 & 52.0 & 47.1 & 49.0\\
 & MLM & 64.6 & 62.1 & 69.9 & 67.1 & 68.4 & 45.8 & 43.3 & 44.3 
& \textbf{56.8} & \textbf{57.2} & 59.9 & 59.5 & 61.6 & \textbf{52.5} & 47.2 & 50.3 \\ \cmidrule(lr){2-18}
 & DS-Seq & 64.1 & \textbf{63.1} & \textbf{70.6} & 67.3 & 68.4 & \textbf{46.0} & 43.4 & 44.2 
& 55.1 & 55.5 & 60.4 & \textbf{60.3} & \textbf{62.8} & 51.1 & 47.3 & 50.3 \\
 & DS-Tok & \textbf{65.0} & 62.9 & 70.1 & \textbf{67.5} & \textbf{68.8} & 45.5 & \textbf{43.9} & \textbf{44.4}
& 54.4 & 55.9 & \textbf{60.9} & 59.8 & 59.7 & 50.2 & \textbf{48.0} & \textbf{50.8} \\
\midrule \midrule
\multirow{4}{*}{\textbf{Average}} & mBERT & 62.4 & 60.4 & 68.4 & 65.7 & 67.0 & 62.7 & 60.5 & 61.4 & 59.8 & 60.8 & 59.3 & 59.4 & 60.2 & 60.2 & 57.2 & 58.0 \\
 & MLM & 63.5 & 62.1 & 69.3 & \textbf{67.0} & 68.0 & 63.4 & 61.3 & \textbf{62.2} & \textbf{62.5} & \textbf{61.7} & 59.8 & 59.4 & 60.5 & \textbf{61.0} & 57.6 & 59.1 \\ \cmidrule(lr){2-18}
 & DS-Seq & 63.3 & \textbf{62.2} & \textbf{69.5} & 66.8 & 68.1 & 63.2 & 60.9 & 62.0 & 60.5 & 60.6 & 60.1 & 60.3 & \textbf{60.8} & 60.9 & \textbf{58.1} & 59.3 \\
 & DS-Tok & \textbf{63.7} & 61.8 & 69.4 & \textbf{67.0} & \textbf{68.6} & \textbf{63.5} & \textbf{61.5} & 62.1 & 60.0 & 60.4 & \textbf{60.5} & \textbf{60.7} & 60.5 & 60.3 & \textbf{58.1} & \textbf{59.4} \\
\bottomrule
\end{tabular}%
}
\vspace{-0.5em}
\caption{Results of gender-specialized (age-specialized) multilingual BERT (DS-Seq and DS-Tok) on gender (age) classification (AC-SA and AC-TD) as intrinsic task and sentiment analysis (SA) and topic detection (TD) as extrinsic evaluation tasks. Comparisons against the vanilla mBERT and mBERT additionally trained on the same review corpora but without demographic information, only with masked language modeling (MLM). For SA and TD, we separately report the performance on the test sets consisting of only one demographic class (gender: F and M, age: <35 and >45) as well as on the mixed test sets containing reviews from both demographic classes (X for both gender and age). Bold numbers indicate the best-performing model (between mBERT, MLM, DS-Seq and DS-Tok) for each country-task combination.}
\vspace{-0.5em}
\label{tab:all_mbert}
\end{table*}
\subsection{Multilingual Specialization Results}
\label{subsec:initial_results}
Table \ref{tab:all_mbert} shows the results of gender- and age-specialized mBERT variants -- DS-Seq and DS-Tok -- on gender and age classification (AC-SA and AC-TD) as intrinsic tasks together with sentiment analysis (SA) and topic detection (TD) as extrinsic evaluation tasks, for each of the five countries encompassed by the Trustpilot datasets~\citep{hovy2015user}. The performance of DS-Seq and DS-Tok is compared against the PLM baselines that have not been exposed to demographic information: vanilla mBERT and mBERT with additional MLM-ing on the same Trustpilot data on which DS-Seq and DS-Tok were trained. 

Our demographically specialized models generally outperform the vanilla mBERT across the board, both on intrinsic and extrinsic tasks, unsurprisingly with much more prominent gains on the former. The comparison against the domain-adaptation in which mBERT was intermediately trained only MLM-ed on Trustpilot reviews, but without demographic category prediction, however, reveals that much of the gains that DS-Seq and DS-Tok have over vanilla mBERT stem from domain adaptation: somewhat surprisingly, DS models fall behind MLM-based domain adaptation on the intrinsic tasks of gender/age classification (e.g., for age group classification on AC-SA, the DS variants fall short of MLM by 2 $F_1$ points), while exhibiting small but fairly consistent gains over MLM for extrinsic SA and TD tasks, both in gender and age intermediate specialization. Although the gains are not particularly convincing, the SA and TD still seem to favor intermediate demographic specialization, which is in line with findings from \citet{hovy-2015-demographic}, who also reported small but (mostly) consistent gains for these two tasks.             
\subsection{Control Experiments}
\label{ssec:control_results}

To more precisely measure the contributions of demographic information that DS-* variants incorporate, we design further experiments that control for two key side-effects of demographic specialization: (i) language specialization and (ii) domain adaptation. We then carry out the meta-regression analysis to tease out the individual contributions of language, domain, and demographic knowledge on the performance difference between vanilla mBERT and respective intermediately specialized models (mBERT or monolingual BERT specialized on the data of the same or different domain with or without demographic signal). Finally, we compare the representations spaces of the PLMs -- before and after demographic specialization -- along the demographic dimension.  


\paragraph{Controlling for Language Proficiency.}
Massively multilingual Transformers (MMTs) like mBERT or XLM-R \cite{conneau-etal-2020-unsupervised} suffer from the \textit{curse of multilinguality} \cite{conneau-etal-2020-unsupervised,lauscher2020zero,pfeiffer2020mad}: given a fixed capacity of the Transformer, the representations from an MMT for any individual (high-resource) language will be of lower quality than those of the monolingual PLM, as MMTs share their limited capacity over many languages. It is thus possible that demographic specialization of mBERT on Trustpilot data in our four languages leads to substantial gains over vanilla mBERT (pretrained on 104 languages) primarily because of mBERT's acquisition of additional language competencies for these four languages. 

To test this, we additionally execute demographic specialization individually for each language (i.e., as opposed to a single multilingual specialization), starting from a monolingual PLM of that language\footnote{We use the following monolingual PLMs from HuggingFace: \texttt{bert-base-cased}, \texttt{bert-base-german-cased}, \texttt{dbmdz/bert-base-french-europeana-cased} and \texttt{Maltehb/danish-bert-botxo}.}. Monolingual PLMs produce higher quality representations for their respective language than mBERT. Because of this, we hypothesize that subjecting them to demographic specialization on Trustpilot is unlikely to improve their ``command'' of the language substantially. Consequently, should we still see (downstream) gains from demographic specialization for monolingual PLMs, we can be more confident that they stem from the injected demographic information.


\setlength{\tabcolsep}{5.7pt}
\begin{table}[t!]
\centering
\scriptsize{
\begin{tabular}{cl ccc ccc}
\toprule
& & \multicolumn{3}{c}{\textit{\textbf{Gender}}} & \multicolumn{3}{c}{\textit{\textbf{Age}}} \\ \cmidrule(lr){3-5}\cmidrule(lr){6-8}
\textbf{Country} & \textbf{Model} & \textbf{AC} & \textbf{SA} & \textbf{TD} & \textbf{AC} & \textbf{SA} & \textbf{TD} \\ \midrule
\multirow{4}{*}{\textbf{Denmark}} & BERT & 65.0 & 70.4 & 59.9 & 66.5 & 66.0 & 56.3 \\ 
& MLM & 65.1 & 70.3 & 60.6 & \textbf{67.4} & \textbf{67.6} & \textbf{57.6} \\ \cmidrule(lr){2-8}
& DS-Seq & \textbf{65.2} & \textbf{70.6} & 60.0 & 67.1 & 67.1 & 56.5\\
& DS-Tok & 65.1 & \textbf{70.6} & \textbf{60.8} & 67.2 & 67.2 & 56.7\\ \midrule
\multirow{4}{*}{\textbf{Germany}} & BERT & 59.4 & 64.3 & 67.8 & 58.8 & 57.1 & 58.3 \\ 
& MLM & \textbf{60.9} & 65.4 & 67.7 & \textbf{60.1} & \textbf{58.1} & \textbf{59.9} \\ \cmidrule(lr){2-8}
& DS-Seq & 60.1 & \textbf{66.2} & 67.8 & 59.8 & 55.8 & 59.1  \\
& DS-Tok & 60.6 & 66.0 & \textbf{67.9} & 58.9 & 54.0 & 59.2 \\ \midrule
\multirow{4}{*}{\textbf{US}} & BERT & 61.5 & 67.1 & 71.0 & 64.1 & 57.2 & \textbf{67.2} \\ 
& MLM & 61.7 & 67.8 & 71.3 & 64.1 & \textbf{60.4} & 66.7 \\ \cmidrule(lr){2-8}
& DS-Seq & 61.6 & \textbf{68.0} & \textbf{71.6} & \textbf{65.2} & 59.4 & 67.1 \\
& DS-Tok & \textbf{62.1} & 67.9 & \textbf{71.6} & 64.3 & 59.4 & 66.7 \\ \midrule
\multirow{4}{*}{\textbf{UK}} & BERT & 64.1 & 72.3 & 70.1 & 65.8 & 65.5 & 68.0 \\ 
& MLM & \textbf{64.3} & \textbf{72.6} & 70.0 & \textbf{66.5} & 66.9 & \textbf{70.0} \\ \cmidrule(lr){2-8}
& DS-Seq & 64.2 & 72.4 & 70.2 & 65.9 & \textbf{67.6} & 69.4 \\
& DS-Tok & 64.1 & 72.2 & \textbf{70.3} & 66.0 & 67.1 & 69.2 \\ \midrule
\multirow{4}{*}{\textbf{France}} & BERT & 63.6 & 68.6 & 45.1 & \textbf{56.5} & 60.3 & 49.6 \\ 
& MLM & \textbf{64.1} & 67.6 & 45.5 & 56.4 & 61.6 & 50.2 \\ \cmidrule(lr){2-8}
& DS-Seq & 63.7 & 69.3 & 45.3 & 56.1 & \textbf{62.0} & 50.2 \\
& DS-Tok & 63.7 & \textbf{69.5} & \textbf{45.6} & 56.3 & 61.5 & \textbf{50.3} \\ \midrule \midrule
\multirow{4}{*}{\textbf{Average}} & BERT & 62.7 & 68.5 & 62.8 & 62.3 & 61.2 & 59.9 \\ 
& MLM & \textbf{63.2} & 68.7 & 63.0 & \textbf{62.9} & \textbf{62.9} & \textbf{60.9} \\ \cmidrule(lr){2-8}
& DS-Seq & 62.9 & \textbf{69.3} & 63.0 & 62.8 & 62.4 & 60.5 \\
& DS-Tok & 63.1 & 69.2 & \textbf{63.2} & 62.5 & 61.8 & 60.4 \\ 
\bottomrule
\end{tabular}%
}
\caption{Results of gender/age-specialized \textbf{monolingual} PLMs -- DS-Seq and DS-Tok -- on demographic attribute classification (AC), sentiment analysis (SA) and topic detection (TD). Bold numbers indicate the best-performing model (between BERT, MLM, DS-Seq and DS-Tok) for each country-task combination.}
\label{tab:monoberts}
\end{table}

\setlength{\tabcolsep}{2.4pt}
\begin{table}[h!]
\centering
\scriptsize{
\begin{tabular}{llgccccc}
\toprule
\textbf{Task} & \textbf{Selected features}& \textbf{all} & \textbf{-D} & \textbf{-M} & \textbf{-S} & \textbf{-C} & \textbf{-A}\\\midrule
 \multicolumn{8}{l}{\textit{\textbf{gender}}} \\ \midrule
\textbf{AC-SA} & \begin{tabular}[c]{@{}l@{}} US (1.0); Denmark (0.9); \\ MLM (0.9); DS-Tok (0.9); 
\end{tabular} \hspace{0.5em}& 0.51 & - & 0.56 & - & 0.63 & 0.62\\\midrule
\textbf{AC-TD} & \begin{tabular}[c]{@{}l@{}} 
MLM (1.0); Monoling\,(1.0) \\ DS-Tok (0.9); \end{tabular} & 0.51 & - & 0.73 & - & 0.54 & 0.66  \\\midrule
\textbf{SA} &\begin{tabular}[c]{@{}l@{}} France~(1.0); DS-Tok~(1.0); \\Denmark~(0.8); MLM~(0.8); \\ In-domain~(0.6) \end{tabular}  & 0.92 & 0.94 & 0.95 & 0.94 & 0.97 & 0.98  \\\midrule
\textbf{TD} & \begin{tabular}[c]{@{}l@{}}
\textsc{DS-Tok} (0.6); MLM (0.5); \\ In-domain (0.5) 
\end{tabular} & 0.33 & 0.36 & 0.35 & 0.34 & 0.35 & 0.40 \\\midrule \midrule
 \multicolumn{8}{l}{\textit{\textbf{age}}} \\ \midrule
\textbf{AC-SA} & \begin{tabular}[c]{@{}l@{}}Denmark~(3.0); \\ MLM~(1.5); 
Monoling~(0.9) \end{tabular} &1.93 & - & 1.98 & - & 2.31 & 2.02 \\\midrule
\textbf{AC-TD} & \begin{tabular}[c]{@{}l@{}} UK~(2.1); France~(1.4); \\ \textsc{MLM} (0.9); 
\end{tabular}  & 0.68 & - & 0.69 & -  & 1.02 & 0.82 \\\midrule
\textbf{SA} &\begin{tabular}[c]{@{}l@{}}
In-domain~(1.3); \\ DS-Tok~(1.0); MLM~(0.9);\end{tabular}  & 0.96 & 1.03 & 0.97 & 0.97 & 0.98 & 1.03 \\\midrule
\textbf{TD} & \begin{tabular}[c]{@{}l@{}}Denmark~(1.6); <35~(0.7); \\ DS-Seq~(0.6); DS-Tok~(0.6) \end{tabular} & 1.52 & 1.53 & 1.53 & 1.55 & 1.61 & 1.54 \\
\bottomrule
\end{tabular}%
}
\vspace{-0.5em}
\caption{Results of meta-regression analysis. We report the goodness-of-fit (RMSE) results for predicting deltas in downstream performance between specialized models and their respective vanilla PLM. Results reported for three tasks -- intrinsic demographic attribute classification (AC; on datasets AC-SA and AC-TD), Sentiment Analysis (SA), and Topic Detection (TD) with both demographic factors, \textit{gender} and \textit{age}. We compare the results across different feature sets -- for all features (\textbf{all}), and excluding individual features: domain (\textbf{-D}), mono- vs. multilingual (\textbf{-M}), fine-tuning demographic setting (e.g., F vs.\,M vs.\,X for gender; \textbf{-S}), country (\textbf{-C}), and the adaptation approach (i.e., MLM vs.\,DS-Tok vs.\,DS-Seq; \textbf{-A}). For each task, when including all features (column: \textbf{all}), we list the most important features, those with weights $>$ 0.5 (\textit{selected features}). 
}
\vspace{-0.5em}
\label{tab:meta}
\end{table}

Table~\ref{tab:monoberts} shows the effects of demographic specialization on monolingual PLMs of the four languages. For brevity (full results in the Appendix), we average the demographic attribute classification (AC) results from two different test portions from Table \ref{tab:all_mbert} (having labels for different downstream tasks, AC-SA and AC-TD); for extrinsic tasks, SA and TD, we report only the score on demographically balanced test sets (denoted ``X'' in Table \ref{tab:all_mbert}).  
The results show that, when we control for language proficiency (as monolingual PLMs are more proficient in their respective language than mBERT), the downstream gains of demographic specialization (on SA and TD) vanish. The DS-Seq and DS-Tok still retain marginal numeric (statistically insignificant) gains over MLM in gender-based specialization, but they lag behind in age-based specialization. Also, both DS-* variants and MLM display only marginal gains with respect to vanilla monolingual BERT models of the four languages: e.g., in gender-specialization and for SA, DS-Tok has an average advantage of $0.7$ $F_1$ over the non-specialized vanilla monolingual BERTs; compare this to a gain of $1.6$ $F_1$ points that mBERT-based DS-Tok has over vanilla mBERT (Table \ref{tab:all_mbert}). These results question the downstream usefulness of demographic specialization -- suggested by findings from prior work \cite{hovy-2015-demographic} and our results for multilingual PLMs (Table \ref{tab:all_mbert}) -- if one starts from the most proficient PLM for the concrete language at hand, i.e., a monolingual PLM.


\begin{figure*}[th]
	\centering
    \includegraphics[trim={0cm 0cm 0cm 0cm}, clip, width=\textwidth]{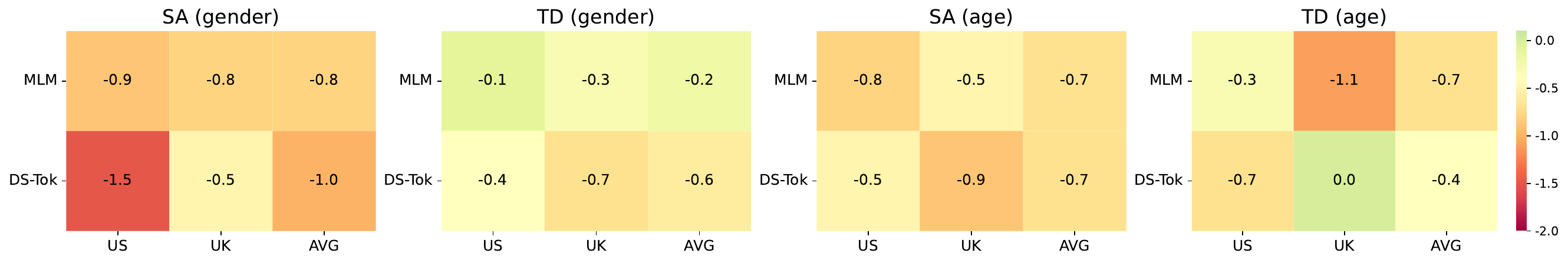}
    
	\caption{Evaluation results on Trustpilot for Sentiment Analysis (SA) and Topic Detection (TD) when running the intermediate specialization on out-of-domain data (RtGender~\citep{voigt-etal-2018-rtgender} for \textit{gender} and BAC~\citep{schler2006effects} for \textit{age}). We report the delta in $F_1$-score in comparison to the specialization on Trustpilot in-domain data.} 
	\label{fig:domain_heatmap}
\end{figure*}

\paragraph{Controlling for Domain Knowledge.}

Both simple additional MLM-ing on Trustpilot data, as well as multi-task demographic specialization training (DS-* variants), inject knowledge about the domain-specific language of reviews into the PLM. As shown by previous work \cite{glavas-etal-2020-xhate,diao2021taming,hung-etal-2022-ds}, domain adaptation generally leads to better downstream performance on in-domain data for any task. We next investigate to which extent the domain specialization is responsible for performance gains. To this end, we perform demographic specialization on (demographically labeled) training data from a different domain: for \emph{gender} specialization, we use the RtGender~\citep{voigt-etal-2018-rtgender} consisting of social media posts collected from diverse sources, 
whereas for \emph{age} specialization we resort to the Blog Authorship Corpus~\citep[BAC;][]{schler2006effects} containing blogposts from \url{blogger.com}.

Figure \ref{fig:domain_heatmap} displays the effects of out-of-domain specialization of BERT on downstream SA and TD performance (i.e., performance differences w.r.t. corresponding in-domain specialized models). Since RtGender and BAC are English-only datasets, we report the results only for US and UK (for brevity, we report the performance only on the demographically balanced test sets, i.e., setups indicated with ``X'' in Table \ref{tab:all_mbert}; both DS-* variants exhibit very similar behavior, so for brevity, we only display results for DS-Tok; complete results are in the Appendix).  
%
Expectedly, the out-of-domain specialization deteriorates the downstream performance for both MLM and DS-Tok. Interestingly, MLM, which is not exposed explicitly to the demographic signal in specialization, tends to suffer less from out-of-domain specialization than the gender-informed DS-Tok. In contrast, age-informed DS-Tok seems to exhibit similar losses as MLM due to out-of-domain specialization. These results further question the hypothesis that demographic information guides downstream gains, suggested by prior work \cite{hovy-2015-demographic} and our in-domain specialization results (with mBERT) from Table \ref{tab:all_mbert}.


\begin{figure*}[t!]
\centering
\begin{subfigure}{\textwidth}
  \includegraphics[trim={1.3cm 1.65cm 0cm 1.95cm},clip,width=\textwidth, ]{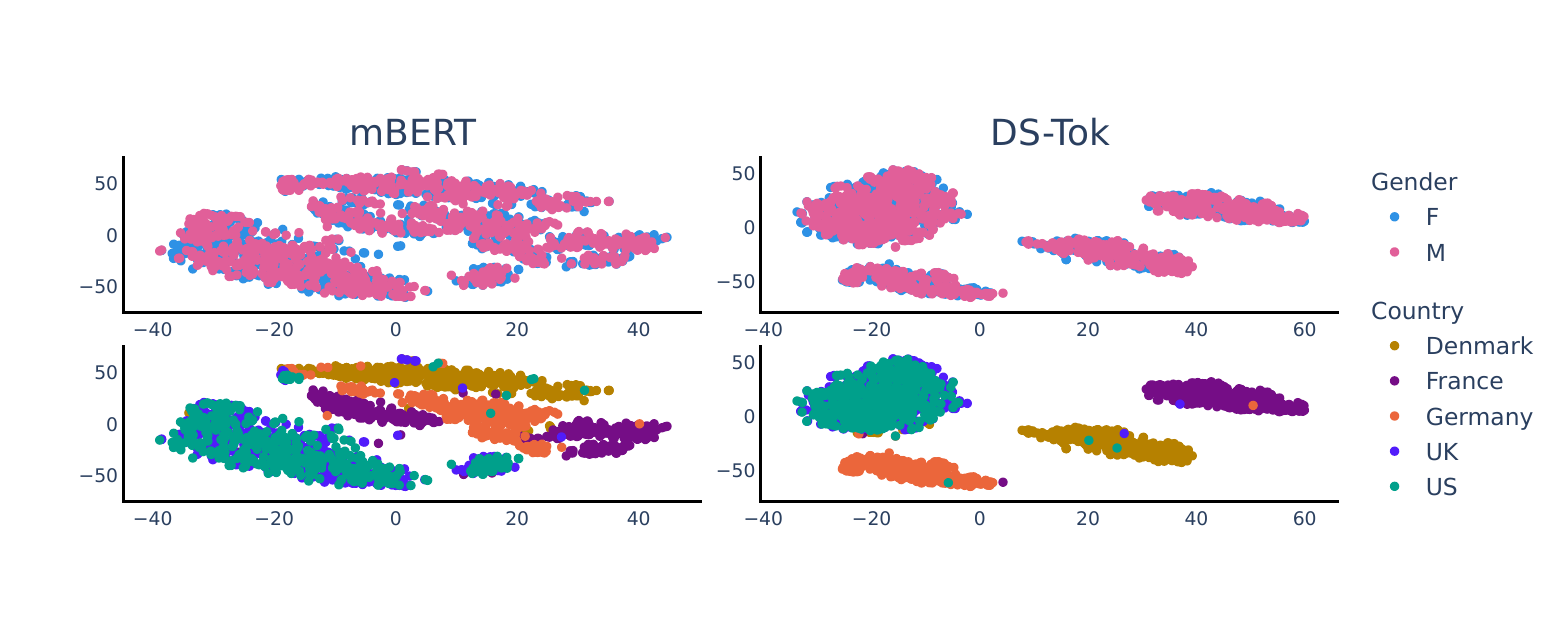}
  \caption{Multilingual}
\end{subfigure} 
\begin{subfigure}{0.465\textwidth}
  \includegraphics[trim={1.3cm 1.65cm 0cm 1.95cm},clip,width=\textwidth]{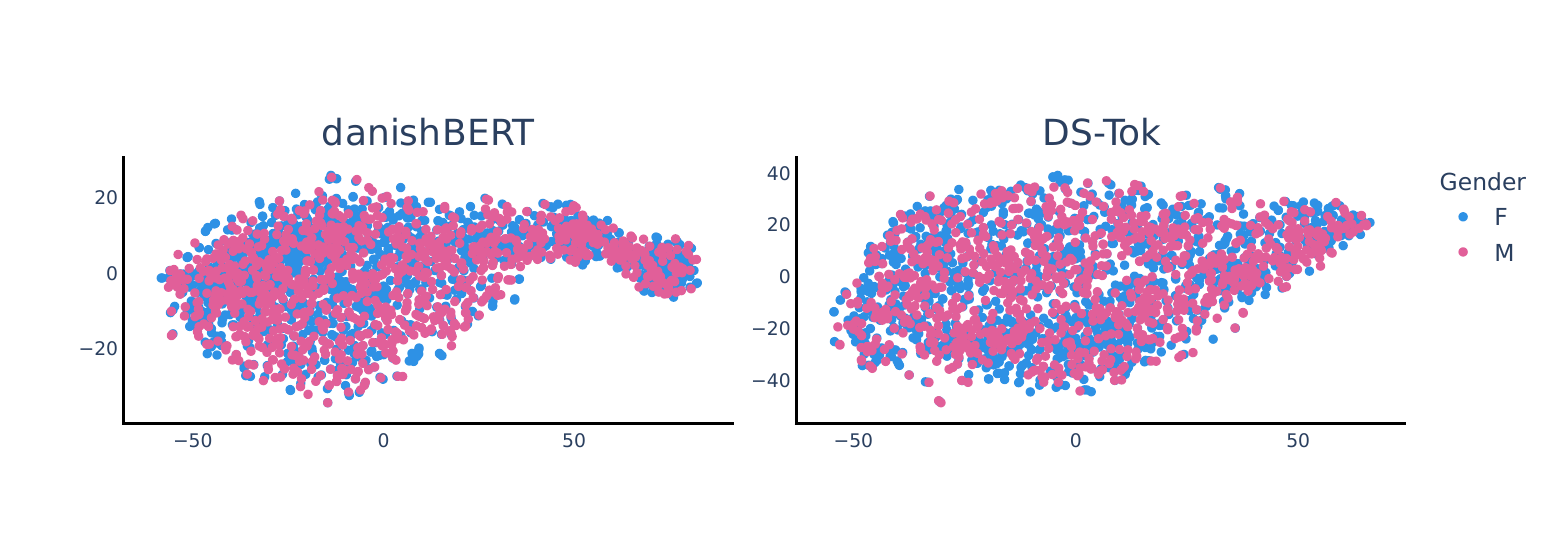}
  \caption{Monolingual (Denmark)}
\end{subfigure}
\quad
\begin{subfigure}{0.465\textwidth}
  \includegraphics[trim={1.2cm 1.65cm 0cm 1.95cm},clip,width=\textwidth]{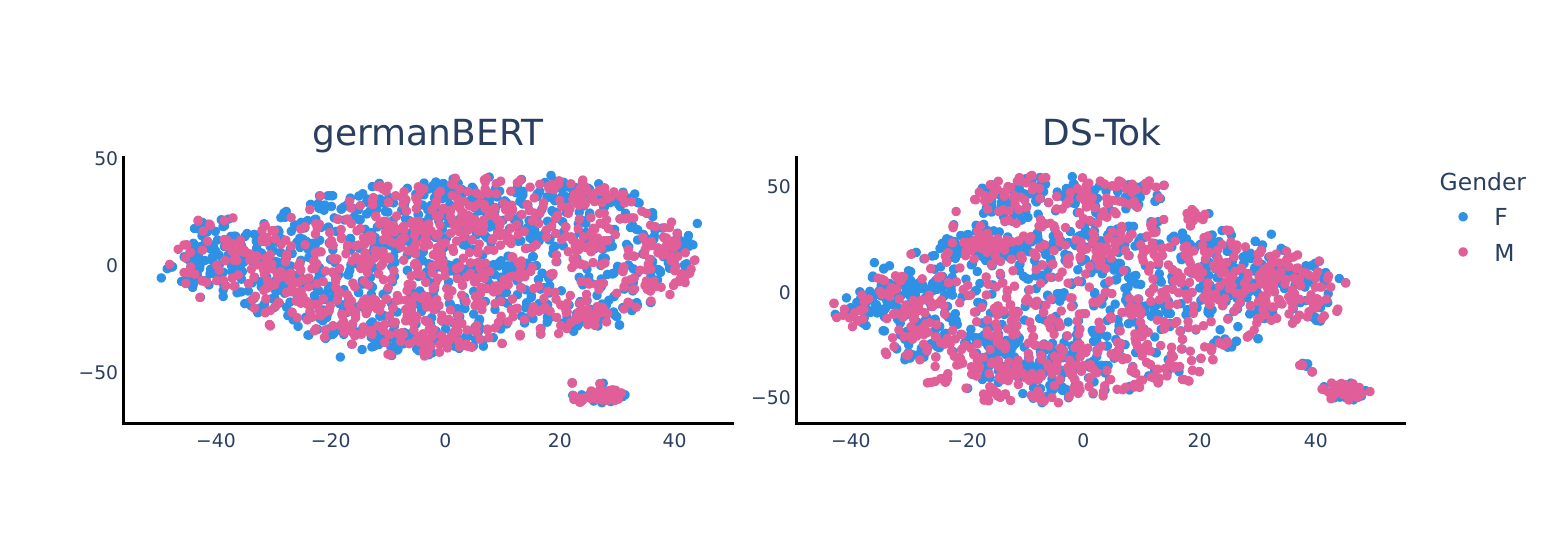}
  \caption{Monolingual (Germany)}
\end{subfigure}
\vspace{-0.5em}
\caption{Results of our multilingual and monolingual qualitative analysis for \emph{gender}. For multilingual case as plotted in (a), we show a tSNE visualization of review texts embedded with a non-specialized (mBERT) and specialized (DS-Tok) model. Colors indicate the demographic subgroup (upper figures) and countries (lower figures), respectively. For monolingual case as illustrated in (b) and (c) for Denmark and Germany, we show a tSNE visualization of texts embedded with
non-specialized (danishBERT, germanBERT) and specialized (DS-Tok) monolingual PLMs. Each subfigure is plotted with 2K instances.}
\vspace{-0.8em}
\label{fig:mbert_ctx_test_all_resize}
\end{figure*}

\vspace{-0.2em}
\paragraph{Meta-regression Analysis.}
Next, we aim to quantify, via a meta-regression analysis, the contributions of individual factors (country, in-domain vs. out-of-domain specialization, language, specialization approach, test set structure) on the task performance (AC-SA, AC-TD, SA, TD). We use the difference in performance between the specialized model and its corresponding vanilla PLM (mBERT or monolingual PLM) as the label (i.e., output, dependent) variable for the regression. We use the following input features (all one-hot encoded) as prediction variables: 
(i) country/language of fine-tuning/evaluation data,
(ii) specialization method (MLM vs. DS-Tok vs. DS-Seq), (iii) in-domain vs. out-of-domain specialization, (iv) whether the starting/vanilla PLM is monolingual (e.g., French BERT) or multilingual (mBERT), (v) and the demographic group from which the fine-tuning/evaluation data comes from (F vs. M vs. X for gender and <35 vs. >45 vs. X for age). We then fit a linear regressor on all data points, using either the full set of features or, in ablations, excluding certain subsets; we report the goodness of fit as average root mean square error (RMSE). 


We summarize the results of our meta-regression analysis in Table~\ref{tab:meta}.
For each task, we list the selected features paired with the RMSE scores.  
When we fit regression using all features (\textbf{all}), the country of origin of fine-tuning data (i.e., features \textit{Denmark}, \textit{France}, \textit{UK}, etc.) tends to overall explain the variance of specialization effect on model performance as good as or even better than the specialization approach (demographically-informed DS-* variants and demographically-uninformed MLM). The specialization approach features (MLM, DS-Tok, and DS-Seq), however, do appear among the most important features in most settings, suggesting that knowing the specialization approach does help predict the performance of the specialized model. Note, however, that in terms of assessing whether demographic information generally improves specialization, this needs to be combined with actual task performance results from Tables \ref{tab:all_mbert} and \ref{tab:monoberts}. For example, feature DS-Tok is among the most important features for SA performance after \textit{gender} specialization: looking at the results for DS-Tok in both Tables \ref{tab:all_mbert} and \ref{tab:monoberts}, we see that it achieves, in most cases, scores above MLM -- this, in turn, suggests that demographically-informed gender specialization does (regardless of other factors) improve the downstream SA performance.      
The ablation results offer a complementary view into the importance of individual features: the larger the increase in RMSE when excluding a feature (compared to using all features), the more important the feature is. The regressions in which we exclude the information on the specialization approach (\textbf{-A}) result in the highest RMSE for gender specialization on both extrinsic tasks (SA and TD). In all other setups (AC for both gender and age specialization, as well as SA and TD for age), there is another type of information, the removal of which results in a less predictable specialization effect: for instance, AC after age specialization, the \textbf{-C} setting increases the RMSE the most, representing that features indicating the demographic composition of the country factor of the fine-tuning dataset jointly have the largest effect on performance. 

Combining results from Tables \ref{tab:all_mbert} and \ref{tab:monoberts} with findings from the meta-regression analysis leads to the overall conclusion that gender-based language specialization of PLMs generally leads to downstream gains, whereas age-based specialization does not. 


\vspace{0.3em}
\noindent\textbf{Qualitative Analysis.}
Finally, we analyze the topology of the PLMs representation space before and after demographic specialization. We encode the reviews from both demographic dimensions -- (i) with the vanilla PLM (mBERT or monolingual BERT) and its DS-Tok specialized counterpart -- and then compress those representations into two dimensions with t-distributed stochastic neighbor embedding~\citep[tSNE;][]{JMLR:v9:vandermaaten08a}. Figure \ref{fig:mbert_ctx_test_all_resize} depicts these representation spaces after gender-specialization (the age-specialization effects lead to similar conclusions; for brevity, we leave them for the Appendix). The tSNE plots do not show any salient gender specialization effect. In the case of mBERT, gender-specialization (corresponding DS-Tok plot) leads to the separation of representation areas according to review language and not gender of its author.\footnote{Note that the green and blue regions, indicating US and UK overlap due to shared language.} In the monolingual cases (illustrated for Danish and German BERT), the space of the gender-specialized encoder visually largely resembles that of the vanilla one, indicating that the demographic specialization procedure (DS-Tok) does not impart dimensions that allow for easy separation of representation space along the specialization dimension (here: gender).

\section{Related Work}

\paragraph{Intermediate Training (Adaptation).}
Intermediate language modeling on texts from the same or similar distribution as the downstream data has been shown to lead to improvements on various NLP tasks~\citep[e.g.,][]{gururangan-etal-2020-dont}. During this process, the goal is to inject additional information into the PLM and thus specialize the model for a particular domain~\citep[e.g.,][]{aharoni-goldberg-2020-unsupervised,hung-etal-2022-ds,BombieriRPF23} or language~\citep[e.g.,][]{glavas-etal-2020-xhate} or to encode other types of knowledge such as common sense knowledge~\citep[e.g.,][]{lauscher-etal-2020-common}, argumentation knowledge~\citep[e.g.,][]{holtermann-etal-2022-fair}, or geographic knowledge~\citep[e.g.,][]{hofmann2022geographic}. 

For instance, \citet{hung-etal-2022-ds} propose a  computationally efficient approach by employing domain-specific adapter modules. They show that their domain adaptation approach leads to improvements in task-oriented dialog. \citet{glavas-etal-2020-xhate} and \citet{hung-etal-2022-multi2woz} perform language adaptation through intermediate MLM in the target languages with filtered text corpora, demonstrating substantial gains in downstream zero-shot cross-lingual transfer for abusive language detection and dialog tasks, respectively. These specialization approaches mainly rely on a single objective (e.g., masked language modeling on ``plain'' text data). Instead, \citet{hofmann2022geographic} conduct geoadaptation by coupling MLM with a token-level geolocation prediction in a dynamic multi-task learning setup. In this work, we adopt a similar approach and perform continued language modeling on the text corpora of a specific demographic dimension. 


\paragraph{Demographic Specialization.}
Language preferences vary with user demographics~\citep{loveys-etal-2018-cross}. Accordingly, several studies have leveraged demographic information (e.g., gender, age, education) to investigate the effect of encoded sociodemographic knowledge in the representations of PLMs~\citep{lauscher-etal-2022-socioprobe} or obtain better language representations for various NLP tasks~\citep{volkova-etal-2013-exploring, garimella-etal-2017-demographic}. Recent research studies on demographic adaptation mainly focus on (1) learning demographic-aware word embeddings and do not work with large PLMs~\citep{hovy-2015-demographic} or (2) leveraging demographic information with special PLM architectures specifically designed for certain downstream tasks (e.g., empathy prediction~\citep{guda-etal-2021-empathbert}). The latter, however, do not consider a task-\textit{agnostic} approach to injecting demographic knowledge into language models, and also focus on a monolingual setup only. Further, what roles the different factors (i.e., domain, language, demographic aspect) in the specialization really play remains unexplored.


\section{Conclusion}

In this work, we thoroughly examined the effects of demographic specialization of Transformers via straightforward injection methods that have been proven effective for other types of knowledge. Initial results on intrinsic and extrinsic evaluation tasks using a multilingual PLM indicated the usefulness of our approach. However, running a series of additional experiments in which we controlled for potentially confounding factors (language and domain) and a meta-analysis indicated that the demographic aspects only have a negligible impact on the downstream  performance. This observation is supported by additional qualitative analysis. Overall, our findings point to the difficulty of injecting demographic knowledge into Transformers: we hope that our in-depth analysis and findings catalyze future research on the topic of truly human-centered NLP, especially in multilingual settings.


\section*{Acknowledgements}

Chia-Chien Hung and Simone Paolo Ponzetto have been supported by the JOIN-T 2 project of the Deutsche Forschungsgemeinschaft (DFG). The work of Anne Lauscher and Dirk Hovy has been funded by the European Research Council (ERC) under the European Union’s Horizon 2020 research and innovation program (grant agreement No.\ 949944, INTEGRATOR).
Anne Lauscher has been additionally supported by the
Excellence Strategy of the German Federal Government and the Länder.
Dirk Hovy is the Scientific director of the Data and Marketing Insights research unit at the Bocconi Institute for Data Science and Analysis.
Goran Glava\v{s} has been supported by the EUINACTION grant from NORFACE Governance (462-19-010, GL950/2-1). We thank the reviewers for their feedback.

\section*{Limitations}
In this paper, we concentrated on the demographic adaptation of PLMs for a few key demographic aspects (i.e., gender and age). There are other known factors, like ethnicity and education, that we cannot explore here. However, there are likely further effects, as well as intersectional effects.
We conducted our experiments using only five Western countries and four Indo-European languages \citep{hovy2015user}, ignoring other world regions and language families. However, due to the scarcity of data, we can only hypothesize that the limited effects of demographic specialization also apply to resource-lean languages (i.e., the language specialization effects are likely to outweigh the ones of the demographic specialization).
Another limitation is the use of pretrained language models, which are all pre-trained on general-purpose data and are freely available. We acknowledge that results may differ for models with greater capacity that have been pretrained on data from other, more specific domains. We primarily concentrate on BERT-like models, which are only a subset of language models, and we leave language model variants for future research.

\section*{Ethics Statement}

Our work deals with demographic adaptation from reviews that should be considered sensitive information. We acknowledge that the limitations in data resources and annotations~\citep{schler2006effects, hovy2015user, voigt-etal-2018-rtgender} give rise to potential risks of overgeneralizing our findings and applying our methods. These risks are due to: (1) \textit{partial language coverage}, where languages are from Indo-European subfamilies that do not represent typologically diverse languages; (2) \textit{limited cultural coverage}~\citep{joshi-etal-2020-state}, where the countries, although speaking different languages, still belong a culturally relatively homogeneous part of the world, i.e., the West; (3) \textit{simplified gender identities}~\citep{dev-etal-2021-harms}, where gender is modeled as a binary variable, which does not reflect the wide variety of possible identities along the gender spectrum and beyond~\citep{lauscher2022welcome}; (4) \textit{unfair stereotypical biases}~\citep{blodgett2020}, namely potential harms that might arise from unfair stereotypical biases in the data (despite our efforts to balance the sample across demographic groups) or pre-encoded in the model \citep{lauscher-etal-2021-sustainable-modular}. Further, the sensitive user profile data might bias the model towards additional demographic characteristics and lead to potentially harmful predictions and applications.

In this work, however, we are interested in advancing NLP research to understand better this fine-grained aspect of the intertwined relationship between demographic adaptation and large pretrained language models in both monolingual and multilingual scenarios. While limited data resources may hinder our ability to fully consider language coverage, cultural coverage, gender identities, and stereotypical biases, it is our obligation to be transparent about these limitations and ethical concerns and to continually work towards improving data collection and methodologies to better serve the needs and perspectives of all users. We believe these insights will lead us toward fairer and more inclusive language technologies. We hope that future research builds on top of our findings and explores other demographic factors, other groups within these factors, and also other languages and countries.



\bibliography{anthology,custom}
\bibliographystyle{acl_natbib_emnlp}

\appendix

\clearpage
\label{sec:appendix}
\onecolumn
\section{Additional Experiments}

\setlength{\tabcolsep}{5.2pt}
\begin{table*}[htp!]
\centering
\scriptsize{
\begin{tabular}{clcccccccccccccccc}
\toprule
\multicolumn{1}{l}{}              & \multicolumn{1}{c}{}         & \multicolumn{4}{c}{\textbf{Gender class.}}  & \multicolumn{6}{c}{\textbf{SA}}                                             & \multicolumn{6}{c}{\textbf{TD}}                                             \\
\cmidrule(lr){3-6}\cmidrule(lr){7-12}\cmidrule(lr){13-18}
\multicolumn{1}{l}{}              &     &  \multicolumn{2}{c}{\textbf{AC-SA}} & \multicolumn{2}{c}{\textbf{AC-TD}}    & F & M & X & F & M & X  & F & M & X  & F & M & X  \\ \cmidrule(lr){3-4}\cmidrule(lr){5-6}
\cmidrule(lr){7-9}\cmidrule(lr){10-12}\cmidrule(lr){13-15}\cmidrule(lr){16-18}
\textbf{Country}               &       \textbf{Model}         & \textbf{Mono}       & \textbf{Multi}      & \textbf{Mono}       & \textbf{Multi}      & \multicolumn{3}{c}{\textbf{Mono}}    & \multicolumn{3}{c}{\textbf{Multi}}   & \multicolumn{3}{c}{\textbf{Mono}}    & \multicolumn{3}{c}{\textbf{Multi}}   \\
\midrule
 \multirow{4}{*}{\textbf{Denmark}} & BERT & 66.1 & 64.0 & 63.8 & 61.8 & 72.3 & 67.9 & 70.4 & 69.2 & 64.8 & 67.2 & 60.7 & 59.8 & 59.9 & 59.3 & 58.3 & 59.0 \\
 & MLM & 66.0 & \textbf{65.2} & \textbf{64.2} & 63.4 & 72.5 & 68.3 & 70.3 & 69.5 & \textbf{65.8} & 67.8 & 60.6 & \textbf{60.6} & 60.6 & 59.7 & 58.8 & \textbf{59.4} \\ \cmidrule(lr){2-18}
 & DS-Seq & \textbf{66.2} & 64.9 & 64.1 & 63.5 & \textbf{72.6} & \textbf{68.6} & \textbf{70.6} & \textbf{69.9} & 65.7 & 67.7 & \textbf{61.3} & 60.5 & 60.0 & 59.7 & 57.8 & 59.1 \\
 & DS-Tok & 66.0 & 65.0 & 64.1 & \textbf{63.5} & 72.4 & 68.4 & 70.6 & 69.1 & 65.6 & \textbf{68.0} & 61.1 & 60.2 & \textbf{60.8} & \textbf{59.9} & \textbf{58.9} & 59.0 \\\midrule
 \multirow{4}{*}{\textbf{Germany}} & BERT & 59.8 & 59.5 & 58.9 & 57.9 & 66.5 & 63.7 & 64.3 & 66.1 & 63.2 & 64.5 & 67.9 & 66.1 & 67.8 & 67.8 & 65.6 & 65.8 \\
 & MLM & 62.0 & 61.2 & 59.7 & 60.1 & 68.1 & \textbf{65.8} & 65.4 & \textbf{67.7} & \textbf{65.3} & 66.1 & 68.5 & 66.7 & 67.7 & \textbf{68.6} & 67.0 & \textbf{67.1} \\\cmidrule(lr){2-18}
 & DS-Seq & \textbf{61.1} & 60.1 & 59.0 & \textbf{60.3} & \textbf{68.8} & 64.4 & \textbf{66.2} & 66.7 & 64.0 & 65.7 & \textbf{68.9} & 66.4 & 67.8 & 67.6 & 65.7 & 66.4 \\
 & DS-Tok & 60.9 & \textbf{62.9} & \textbf{60.3} & 58.3 & 67.9 & 65.6 & 66.0 & 66.8 & 64.3 & \textbf{66.8} & 68.6 & \textbf{66.8} & \textbf{67.9} & 68.3 & \textbf{67.0} & 66.7 \\\midrule
\multirow{4}{*}{\textbf{US}} & BERT & 64.3 & 62.6 & 58.7 & 58.1 & 68.6 & 67.0 & 67.1 & 66.3 & 64.4 & 66.0 & 72.5 & 69.7 & 71.0 & 71.2 & 68.4 & 70.2 \\
 & MLM & 64.6 & 63.3 & 58.7 & \textbf{59.6} & 68.4 & 67.6 & 67.8 & 67.3 & 66.2 & 66.9 & 73.1 & 70.1 & 71.3 & 72.1 & 69.4 & 70.3 \\\cmidrule(lr){2-18}
 & DS-Seq & 64.3 & \textbf{63.8} & 58.8 & 59.2 & 68.6 & \textbf{68.0} & \textbf{68.0} & 67.2 & 66.3 & 67.0 & 73.1 & \textbf{70.3} & 71.6 & 72.3 & 69.2 & 70.4 \\
 & DS-Tok & \textbf{64.7} & 62.2 & \textbf{59.4} & 58.8 & \textbf{68.9} & 67.5 & 67.9 & \textbf{68.0} & \textbf{66.4} & \textbf{67.3} & \textbf{73.3} & 69.9 & \textbf{71.6} & \textbf{72.8} & \textbf{69.5} & \textbf{70.5} \\\midrule
 \multirow{4}{*}{\textbf{UK}} & BERT & 63.2 & 61.9 & 65.0 & 63.1 & 73.4 & 71.0 & 72.3 & 71.0 & 69.0 & 69.7 & 71.2 & 69.1 & 70.1 & 70.4 & 67.9 & 68.9 \\
 & MLM & \textbf{63.7} & 63.0 & 64.8 & 65.3 & \textbf{73.9} & 71.0 & \textbf{72.6} & 72.0 & 70.4 & 71.0 & 71.2 & \textbf{69.4} & 70.0 & 70.6 & 67.9 & 69.8 \\\cmidrule(lr){2-18}
 & DS-Seq & 63.2 & 63.4 & \textbf{65.2} & 64.9 & 73.6 & \textbf{72.2} & 72.4 & 72.9 & 70.9 & 71.7 & \textbf{71.5} & 69.3 & 70.2 & 70.6 & 68.2 & 69.8 \\
 & DS-Tok & 63.3 & \textbf{63.5} & 64.8 & \textbf{65.6} & 73.7 & 72.0 & 72.2 & \textbf{73.0} & \textbf{71.0} & \textbf{71.9} & 71.4 & 69.1 & \textbf{70.3} & \textbf{70.8} & \textbf{68.2} & \textbf{69.9} \\\midrule
 \multirow{4}{*}{\textbf{France}} & BERT & 64.1 & 63.9 & 63.1 & 61.2 & 70.5 & 67.3 & 68.6 & 69.3 & 67.0 & 67.8 & 46.0 & \textbf{44.5} & 45.1 & 44.6 & 42.4 & 43.1 \\
 & MLM & \textbf{64.9} & 64.6 & \textbf{63.2} & 62.1 & 71.0 & 67.7 & 67.6 & 69.9 & 67.1 & 68.4 & 46.2 & 44.3 & 45.5 & 45.8 & 43.3 & 44.3 \\\cmidrule(lr){2-18}
 & DS-Seq & 64.2 & 64.1 & 63.1 & \textbf{63.1} & 70.5 & 67.5 & 69.3 & \textbf{70.6} & 67.3 & 68.4 & \textbf{47.1} & 44.2 & 45.3 & \textbf{46.0} & 43.4 & 44.2 \\
 & DS-Tok & 64.4 & \textbf{65.0} & 62.9 & 62.9 & \textbf{71.7} & \textbf{68.3} & \textbf{69.5} & 70.1 & \textbf{67.5} & \textbf{68.8} & 46.9 & 44.3 & \textbf{45.6} & 45.5 & \textbf{43.9} & \textbf{44.4}\\\midrule \midrule

\multirow{4}{*}{\textbf{Average}} & BERT & 63.5 & 62.4 & 61.9 & 60.4 & 70.3 & 67.4 & 68.5 & 68.4 & 65.7 & 67.0 & 63.7 & 61.8 & 62.8 & 62.7 & 60.5 & 61.4 \\
 & MLM &  \textbf{64.2} & 63.5 & 62.1 & 62.1 & 70.8 & 68.1 & 68.7 & 69.3 & \textbf{67.0} & 68.0 & 63.9 & \textbf{62.2} & 63.0 & 63.4 & 61.3 & \textbf{62.2}\\\cmidrule(lr){2-18}
 & DS-Seq & 63.8 & 63.3 & 62.0 & \textbf{62.2} & 70.8 & 68.1 & \textbf{69.3} & \textbf{69.5} & 66.8 & 68.1 & \textbf{64.4} & 62.1 & 63.0 & 63.2 & 60.9 & 62.0 \\
 & DS-Tok &  63.9 & \textbf{63.7} & \textbf{62.3} & 61.8 & \textbf{70.9} & \textbf{68.4} & 69.2 & 69.4 & \textbf{67.0} & \textbf{68.6} & 64.3 & 62.1 & \textbf{63.2} & \textbf{63.5} & \textbf{61.5} & 62.1\\\bottomrule
 
\end{tabular}%
}
\vspace{-0.3em}
\caption{Evaluation results compared with monolingual BERT and multilingual BERT (mBERT) on five countries with \textit{gender} data for intrinsic attribute classification tasks (AC-SA, AC-TD) and extrinsic evaluation tasks: sentiment analysis (SA) and topic detection (TD).}
\label{tab:mono_multi_compare_gender}
\end{table*}

\setlength{\tabcolsep}{5.2pt}
\begin{table*}[t]
\centering
\vspace{-0.3em}
\scriptsize{
\begin{tabular}{clcccccccccccccccc}
\toprule
\multicolumn{1}{l}{}              & \multicolumn{1}{c}{}         & \multicolumn{4}{c}{\textbf{Age class.}} & \multicolumn{6}{c}{\textbf{SA}}                                             & \multicolumn{6}{c}{\textbf{TD}}                                             \\
\cmidrule(lr){3-6}\cmidrule(lr){7-12}\cmidrule(lr){13-18}
\multicolumn{1}{l}{}              &                              &            \multicolumn{2}{c}{\textbf{AC-SA}} & \multicolumn{2}{c}{\textbf{AC-TD}}      & <35 & >45 & X & <35 & >45 & X & <35 & >45 & X & <35 & >45 & X \\
\cmidrule(lr){3-4}\cmidrule(lr){5-6}\cmidrule(lr){7-9}\cmidrule(lr){10-12}\cmidrule(lr){13-15}\cmidrule(lr){16-18}
\multicolumn{1}{l}{\textbf{Country}}              &       \textbf{Model}         & \textbf{Mono}       & \textbf{Multi}      & \textbf{Mono}       & \textbf{Multi}      & \multicolumn{3}{c}{\textbf{Mono}}    & \multicolumn{3}{c}{\textbf{Multi}}   & \multicolumn{3}{c}{\textbf{Mono}}    & \multicolumn{3}{c}{\textbf{Multi}}   \\
\midrule
 \multirow{4}{*}{\textbf{Denmark}} & BERT & 67.7 & 57.2 & 65.3 & 64.5 & 67.3 & 66.2 & 66.0 & 62.7 & 62.7 & 62.9 & 58.4 & 54.4 & 56.3 & 56.1 & 52.2 & 53.4 \\
 & MLM & 67.4 & \textbf{65.5} & \textbf{67.4} & 65.1 & \textbf{67.7} & \textbf{67.3} & \textbf{67.6} & 63.3 & 62.1 & 63.0 & \textbf{59.3} & 55.3 & \textbf{57.6} & \textbf{57.1} & 52.6 & 54.1 \\\cmidrule(lr){2-18}
 & DS-Seq & 67.4 & 65.2 & 66.8 & \textbf{65.2} & 67.4 & 66.2 & 67.1 & 63.1 & 62.9 & 63.0 & 58.7 & 55.0 & 56.5 & 56.9 & \textbf{53.3} & \textbf{54.5} \\
 & DS-Tok & \textbf{67.8} & 65.3 & 66.6 & 64.6 & 67.6 & 66.1 & 67.2 & \textbf{64.2} & \textbf{63.3} & \textbf{63.2} & 59.0 & \textbf{55.4} & 56.7 & 56.2 & 53.2 & 54.3 \\\midrule
 \multirow{4}{*}{\textbf{Germany}} & BERT & 57.9 & 58.0 & 59.6 & 56.9 & 53.6 & 57.9 & 57.1 & 52.6 & 55.0 & 55.0 & 61.6 & 57.4 & 58.3 & 60.1 & 55.3 & 57.1 \\
 & MLM & 58.1 & \textbf{61.1} & \textbf{62.0} & \textbf{58.9} & \textbf{58.1} & \textbf{58.2} & \textbf{58.1} & 53.6 & 55.5 & 56.7 & 62.2 & 57.6 & \textbf{59.9} & \textbf{61.5} & 56.5 & 58.7 \\\cmidrule(lr){2-18}
 & DS-Seq & \textbf{58.2} & 56.4 & 61.3 & 58.2 & 56.3 & 57.3 & 55.8 & \textbf{53.8} & 55.3 & 55.5 & 63.5 & 57.9 & 59.1 & 60.8 & \textbf{57.6} & \textbf{59.3} \\
 & DS-Tok & 57.2 & 56.6 & 60.6 & 57.4 & 57.9 & 58.1 & 54.0 & 53.0 & \textbf{56.5} & \textbf{56.7} & \textbf{63.5} & \textbf{58.2} & 59.2 & 59.3 & 56.5 & 59.3 \\\midrule
 \multirow{4}{*}{\textbf{US}} & BERT & 65.2 & 62.9 & 63.0 & 60.7 & 60.5 & 58.7 & 57.2 & 57.7 & 57.9 & 57.8 & 68.8 & 64.9 & \textbf{67.2} & 68.0 & 64.3 & 64.3 \\
 & MLM & 65.3 & \textbf{63.6} & 62.9 & \textbf{61.9} & 59.8 & \textbf{59.5} & \textbf{60.4} & 59.4 & 57.8 & \textbf{58.2} & 71.2 & 65.7 & 66.7 & 69.0 & 64.2 & 65.2 \\\cmidrule(lr){2-18}
 & DS-Seq & \textbf{66.2} & 60.7 & \textbf{64.1} & 61.5 & \textbf{61.6} & 58.3 & 59.4 & 59.3 & 57.9 & 58.0 & \textbf{72.5} & 65.5 & 67.1 & \textbf{69.8} & 64.4 & \textbf{65.8} \\
 & DS-Tok & 65.7 & 59.7 & 62.9 & 61.2 & 61.1 & 58.7 & 59.4 & \textbf{59.9} & \textbf{58.6} & 57.8 & 69.4 & \textbf{65.7} & 66.7 & 69.2 & \textbf{65.4} & 64.9 \\\midrule
\multirow{4}{*}{\textbf{UK}} & BERT & 65.7 & 65.1 & 65.8 & 65.2 & 65.2 & 66.3 & 65.5 & 63.8 & 63.9 & 63.7 & 68.1 & 68.1 & 68.0 & 64.7 & 67.1 & 66.3 \\
 & MLM & 66.9 & \textbf{65.4} & \textbf{66.1} & \textbf{65.6} & \textbf{68.2} & \textbf{67.2} & 66.9 & 62.8 & 62.0 & 63.0 & \textbf{68.8} & \textbf{70.1} & \textbf{70.0} & 65.1 & 67.3 & 67.3 \\\cmidrule(lr){2-18}
 & DS-Seq & 67.0 & 65.3 & 64.7 & 62.8 & 67.8 & 66.4 & \textbf{67.6} & 63.8 & 64.9 & 64.9 & 67.8 & 68.9 & 69.4 & 66.0 & \textbf{68.1} & 66.5 \\
 & DS-Tok & \textbf{66.8} & 64.0 & 65.2 & 62.8 & 67.6 & 66.5 & 67.1 & \textbf{64.6} & \textbf{65.2} & \textbf{65.1} & 68.2 & 69.6 & 69.2 & \textbf{66.4} & 67.3 & \textbf{67.6} \\\midrule
 \multirow{4}{*}{\textbf{France}} & BERT & \textbf{56.0} & 55.7 & \textbf{57.0} & 56.6 & 59.7 & 57.5 & 60.3 & 59.6 & 57.4 & 61.5 & 51.9 & 49.1 & 49.6 & 52.0 & 47.1 & 49.0 \\
 & MLM & 55.9 & \textbf{56.8} & 56.9 & \textbf{57.2} & 60.7 & 59.4 & 61.6 & 59.9 & 59.5 & 61.6 & 53.8 & 48.5 & 50.2 & \textbf{52.5} & 47.2 & 50.3 \\\cmidrule(lr){2-18}
 & DS-Seq & 55.5 & 55.1 & 56.7 & 55.5 & \textbf{61.3} & 58.7 & \textbf{62.0} & 60.4 & \textbf{60.3} & \textbf{62.8} & 53.8 & 49.0 & 50.2 & 51.1 & 47.3 & 50.3 \\
 & DS-Tok & 55.8 & 54.4 & 56.7 & 55.9 & 60.2 & \textbf{60.7} & 61.5 & \textbf{60.9} & 59.8 & 59.7 & \textbf{54.6} & \textbf{51.4} & \textbf{50.3} & 50.2 & \textbf{48.0} & \textbf{50.8}
 \\ \midrule \midrule
\multirow{4}{*}{\textbf{Average}} & BERT & 62.5 & 59.8 & 62.1 & 60.8 & 61.3 & 61.3 & 61.2 & 59.3 & 59.4 & 60.2 & 61.8 & 58.8 & 59.9 & 60.2 & 57.2 & 58.0 \\
 & MLM & 62.7 & \textbf{62.5} & \textbf{63.1} & \textbf{61.7} & \textbf{62.9} & \textbf{62.3} & \textbf{62.9} & 59.8 & 59.4 & 60.5 & 63.1 & 59.4 & \textbf{60.9} & \textbf{61.0} & 57.6 & 59.1 \\\cmidrule(lr){2-18}
 & DS-Seq & \textbf{62.9} & 60.5 & 62.7 & 60.6 & \textbf{62.9} & 61.4 & 62.4 & 60.1 & 60.3 & \textbf{60.8} & \textbf{63.3} & 59.3 & 60.5 & 60.9 & \textbf{58.1} & 59.3 \\
 & DS-Tok & 62.7 & 60.0 & 62.4 & 60.4 & \textbf{62.9} & 62.0 & 61.8 & \textbf{60.5} & \textbf{60.7} & 60.5 & 62.9 & \textbf{60.1} & 60.4 & 60.3 & \textbf{58.1} & \textbf{59.4} \\
\bottomrule
\end{tabular}%
}
\vspace{-0.3em}
\caption{Evaluation results compared with monolingual BERT and multilingual BERT (mBERT) on five countries with \textit{age} data for intrinsic attribute classification tasks (AC-SA, AC-TD) and extrinsic evaluation tasks: sentiment analysis (SA) and topic detection (TD).}
\label{tab:mono_multi_compare_age}
\end{table*}

\setlength{\tabcolsep}{9.2pt}
\begin{table*}[t]
\centering
\scriptsize{%
\begin{tabular}{llcccccccccccc}
\toprule
 &  & \multicolumn{6}{c}{\textbf{SA}} & \multicolumn{6}{c}{\textbf{TD}} \\\cmidrule(lr){3-8}\cmidrule(lr){9-14}
 \textit{\textbf{gender}} &  & F & M & X & F & M & X & F & M & X & F & M & X \\\cmidrule(lr){3-5}\cmidrule(lr){6-8}\cmidrule(lr){9-11}\cmidrule(lr){12-14}
  \textbf{Country} & \textbf{Model} & \multicolumn{3}{c}{\textbf{RtGender}} & \multicolumn{3}{c}{\textbf{Trustpilot}} & \multicolumn{3}{c}{\textbf{RtGender}} & \multicolumn{3}{c}{\textbf{Trustpilot}} \\\midrule
\multirow{3}{*}{\textbf{US}} & MLM & 68.3 & 67.3 & 66.9 & 68.4 & 67.6 & 67.8 & \textbf{72.7} & \textbf{69.9} & 71.1 & 73.1 & 70.1 & 71.3 \\ \cmidrule(lr){2-14}
 & DS-Seq & 68.1 & \textbf{67.4} & \textbf{66.9} & 68.6 & \textbf{68.0} & \textbf{68.0} & 72.7 & 69.3 & \textbf{71.2} & 73.1 & \textbf{70.3} & 71.6 \\
 & DS-Tok & \textbf{68.6} & 67.2 & 66.4 & \textbf{68.9} & 67.5 & 67.9 & 72.4 & 69.6 & 71.2 & \textbf{73.3} & 69.9 & \textbf{71.6} \\\midrule
\multirow{3}{*}{\textbf{UK}} & MLM & 73.3 & 71.0 & 71.7 & \textbf{73.9} & 71.0 & \textbf{72.6} & 71.1 & \textbf{69.3} & \textbf{69.8} & 71.2 & \textbf{69.4} & 70.0 \\\cmidrule(lr){2-14}
 & DS-Seq & 73.3 & 71.1 & \textbf{71.9} & 73.6 & \textbf{72.2} & 72.4 & 71.2 & 69.0 & 69.5 & \textbf{71.5} & 69.3 & 70.2 \\
 & DS-Tok & \textbf{73.4} & \textbf{71.1} & 71.6 & 73.7 & 72.0 & 72.2 & \textbf{71.3} & 69.2 & 69.6 & 71.4 & 69.1 & \textbf{70.3} \\\midrule\midrule
 \multirow{3}{*}{\textbf{Average}} & MLM & 70.8 & 69.2 & 69.3 & 71.2 & 69.3 & \textbf{70.2} & 71.9 & \textbf{69.6} & \textbf{70.5} & 72.2 & \textbf{69.8} & 70.7 \\\cmidrule(lr){2-14}
 & DS-Seq & 70.7 & \textbf{69.3} & \textbf{69.4} & 71.1 & \textbf{70.1} & \textbf{70.2} & \textbf{72.0} & 69.2 & 70.4 & 72.3 & \textbf{69.8} & 70.9 \\
 & DS-Tok & \textbf{71.0} & 69.2 & 69.0 & \textbf{71.3} & 69.8 & 70.1 & 71.9 & 69.4 & 70.4 & \textbf{72.4} & 69.5 & \textbf{71.0} \\\midrule
 & \multicolumn{1}{c}{} & \multicolumn{6}{c}{\textbf{SA}} & \multicolumn{6}{c}{\textbf{TD}} \\\cmidrule(lr){3-8}\cmidrule(lr){9-14}
\textit{\textbf{age}} & \multicolumn{1}{c}{} & <35 & >45 & X  & <35 & >45 & X  & <35 & >45 & X  & <35 & >45 & X  \\\cmidrule(lr){3-5}\cmidrule(lr){6-8}\cmidrule(lr){9-11}\cmidrule(lr){12-14}
\textbf{Country} & \textbf{Model} & \multicolumn{3}{c}{\textbf{BAC}} & \multicolumn{3}{c}{\textbf{Trustpilot}} & \multicolumn{3}{c}{\textbf{BAC}} & \multicolumn{3}{c}{\textbf{Trustpilot}} \\\midrule
\multirow{3}{*}{\textbf{US}} & MLM & \textbf{59.4} & 58.4 & \textbf{58.9} & 59.8 & \textbf{59.5} & \textbf{60.4} & 68.4 & 64.6 & 66.9 & 71.2 & 65.7 & 66.7 \\\cmidrule(lr){2-14}
 & DS-Seq & 58.4 & 57.3 & 58.0 & \textbf{61.6} & 58.3 & 59.4 & 68.6 & 64.5 & \textbf{67.3} & \textbf{72.5} & 65.5 & \textbf{67.1} \\
 & DS-Tok & 58.6 & \textbf{58.5} & 58.9 & 61.1 & 58.7 & 59.4 & \textbf{69.3} & \textbf{65.0} & 67.1 & 69.4 & \textbf{65.7} & 66.7 \\\midrule
\multirow{3}{*}{\textbf{UK}} & MLM & 66.2 & \textbf{66.7} & 66.4 & \textbf{68.2} & \textbf{67.2} & 66.9 & 67.8 & 68.7 & 68.9 & \textbf{68.8} & \textbf{70.1} & \textbf{70.0} \\\cmidrule(lr){2-14}
 & DS-Seq & 66.1 & 66.6 & \textbf{66.8} & 67.8 & 66.4 & \textbf{67.6} & 67.8 & 68.7 & 68.6 & 67.8 & 68.9 & 69.4 \\
 & DS-Tok & \textbf{66.6} & 66.0 & 66.3 & 67.6 & 66.5 & 67.1 & \textbf{68.0} & \textbf{68.8} & \textbf{69.2} & 68.2 & 69.6 & 69.2 \\
 \midrule\midrule
 \multirow{3}{*}{\textbf{Average}} & MLM & \textbf{62.8} & \textbf{62.6} & \textbf{62.7} & 64.0 & \textbf{63.4} & \textbf{63.7} & 68.1 & 66.7 & 67.9 & 70.0 & \textbf{67.9} & \textbf{68.4} \\\cmidrule(lr){2-14}
 & DS-Seq & 62.3 & 62.0 & 62.4 & \textbf{64.7} & 62.4 & 63.5 & 68.2 & 66.6 & 68.0 & \textbf{70.2} & 67.2 & 68.3 \\
 & DS-Tok & 62.6 & 62.3 & 62.6 & 64.4 & 62.6 & 63.3 & \textbf{68.7} & \textbf{66.9} & \textbf{68.2} & 68.8 & 67.7 & 68.0 \\
 \bottomrule
\end{tabular}%
}
\caption{Evaluation results on Trustpilot classification tasks (SA, TD) compared by specializing on out-domain data (RtGender~\citep{voigt-etal-2018-rtgender} for \textit{gender} and BAC~\citep{schler2006effects} for \textit{age}) and in-domain data (Trustpilot).}
\label{tab:domain_compare}
\end{table*}

\begin{figure*}[p!]
\centering
\begin{subfigure}{\textwidth}
  \includegraphics[trim={1.3cm 1.65cm 0cm 1.95cm},clip,width=\textwidth]{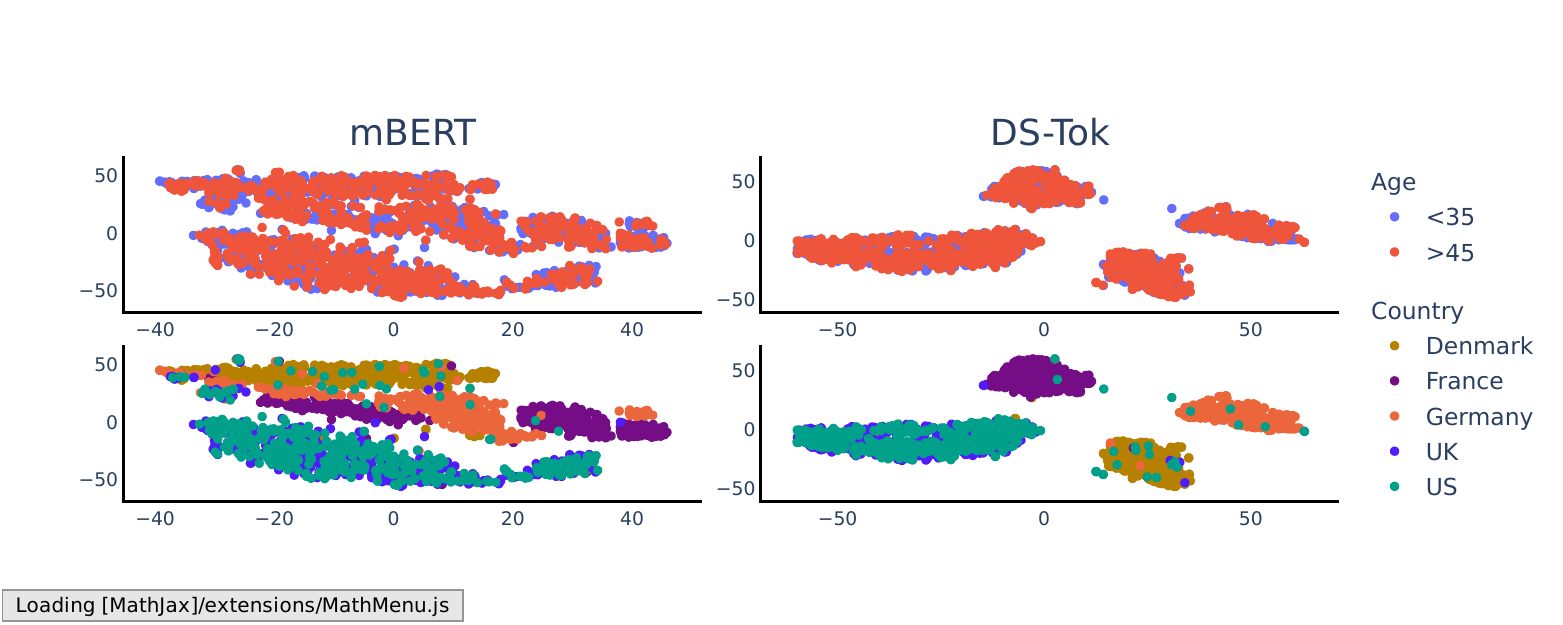}
  \caption{Multilingual}
\end{subfigure}
\begin{subfigure}{0.465\textwidth}
  \includegraphics[trim={1.3cm 1.65cm 0cm 1.95cm},clip,width=\textwidth]{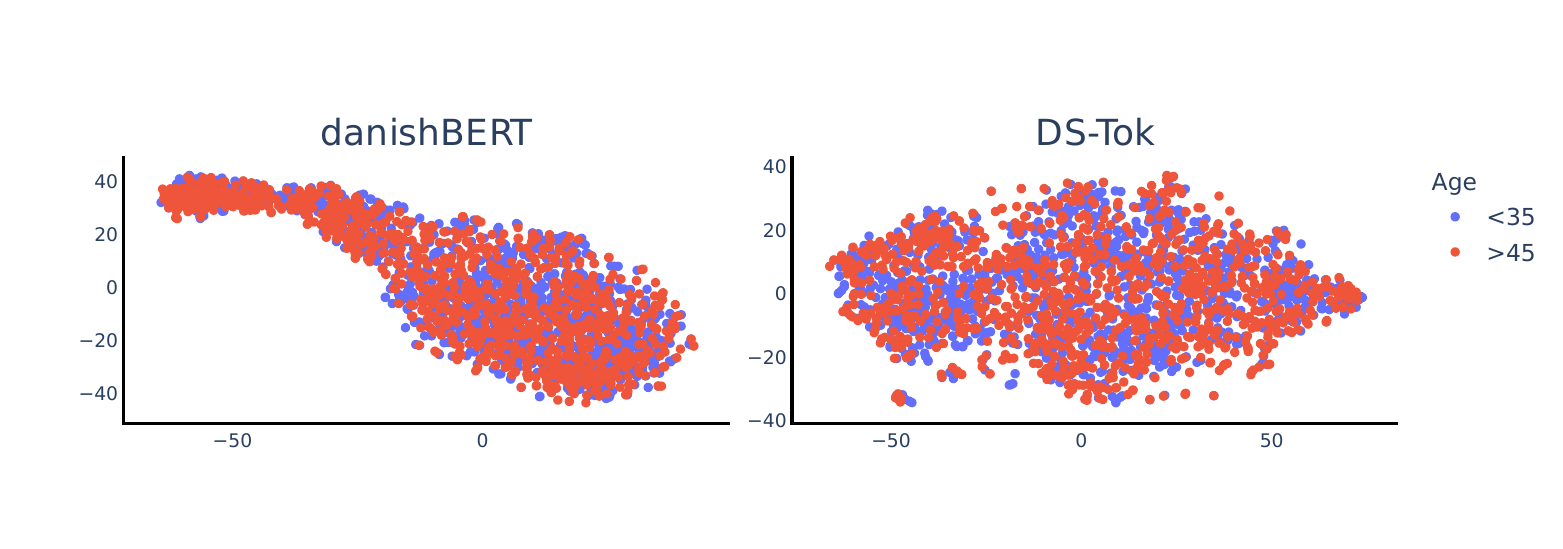}
  \caption{Monolingual (Denmark)}
\end{subfigure}
\quad
\begin{subfigure}{0.465\textwidth}
  \includegraphics[trim={1.2cm 1.65cm 0cm 1.95cm},clip,width=\textwidth]{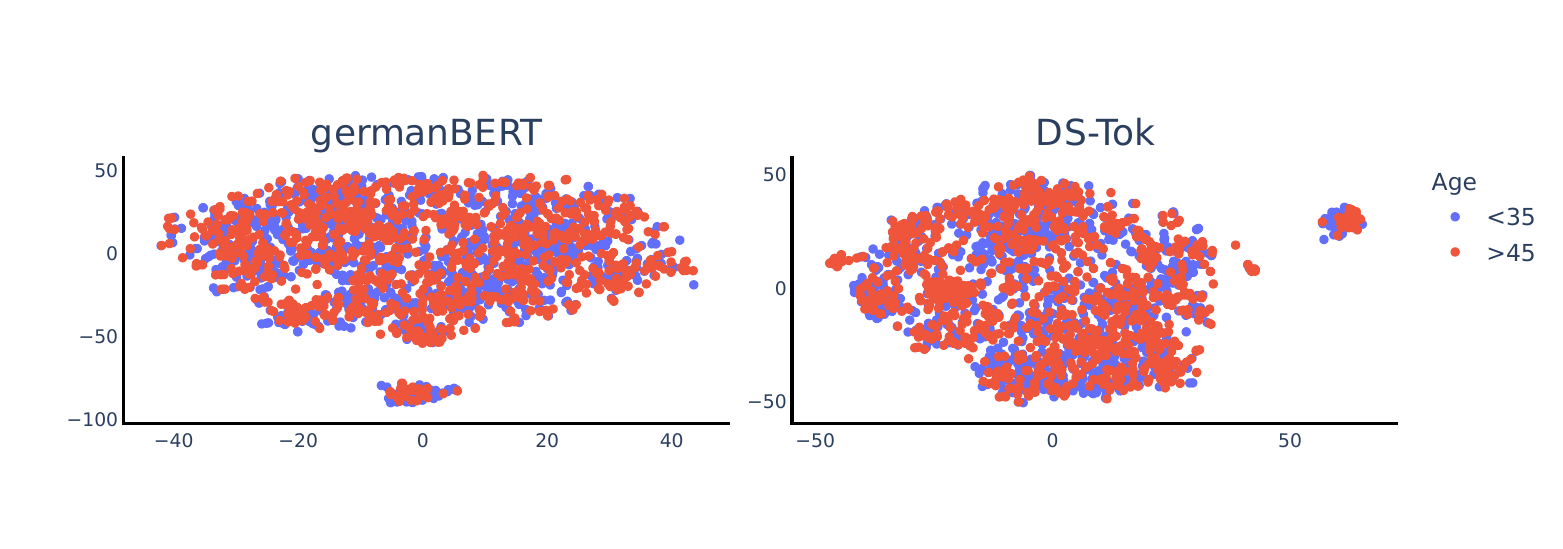}
  \caption{Monolingual (Germany)}
\end{subfigure}
\vspace{-0.7em}
\caption{Results of our multilingual and monolingual qualitative analysis for \emph{age}. For multilingual case as plotted in (a), we show a tSNE visualization of review texts embedded with a non-specialized (mBERT) and specialized (DS-Tok) model. Colors indicate the demographic subgroup (upper figures) and countries (lower figures), respectively. For monolingual case as illustrated in (b) and (c) for Denmark and Germany, we show a tSNE visualization of texts embedded with
non-specialized (danishBERT, germanBERT) and specialized (DS-Tok) monolingual PLMs. Each subfigure is plotted with 2K instances.}
\vspace{-0.7em}
\label{fig:monobert_ctx_test}
\end{figure*}

\end{document}